%% file: Main.tex
\documentclass[11pt]{article}

\usepackage[final]{acl}

\usepackage{times}
\usepackage{latexsym}

\usepackage[T1]{fontenc}
\usepackage[utf8]{inputenc}

\usepackage{microtype}
\usepackage{inconsolata}
\usepackage{graphicx}

\usepackage{float}
\usepackage{times}
\usepackage{latexsym}
\usepackage{epigraph} 
\usepackage[most]{tcolorbox}
\usepackage{enumitem}
\usepackage[rgb]{xcolor}
\usepackage{booktabs}
\usepackage{multirow}
\usepackage{pifont}
\usepackage{xspace}
\usepackage{tabularx}
\usepackage{wrapfig}
\usepackage[normalem]{ulem}
\useunder{\uline}{\ul}{}
\usepackage{booktabs}
\usepackage{multirow}
\usepackage{pifont}
\usepackage{orcidlink}
\usepackage{amsmath}
\usepackage{amssymb}
\usepackage{threeparttable}
\usepackage{makecell}
\usepackage{enumitem}
\usepackage{graphicx}
\usepackage[table,xcdraw]{xcolor}
\usepackage{colortbl}
\usepackage{float}
\usepackage{placeins}

\definecolor{codegreen}{RGB}{93, 168, 128} 
\definecolor{codeblue}{RGB}{74, 74, 250}  
\definecolor{codered}{RGB}{209, 106, 114}
\definecolor{text}{RGB}{68,114,157}

\usepackage{listings}
\newcommand{\blackcircle}[1]{%
    \raisebox{0.8pt}{\textcircled{\raisebox{-0.8pt}{\small #1}}}
}

\newcommand{\SpiritRAG}{\includegraphics[width=1.8em]{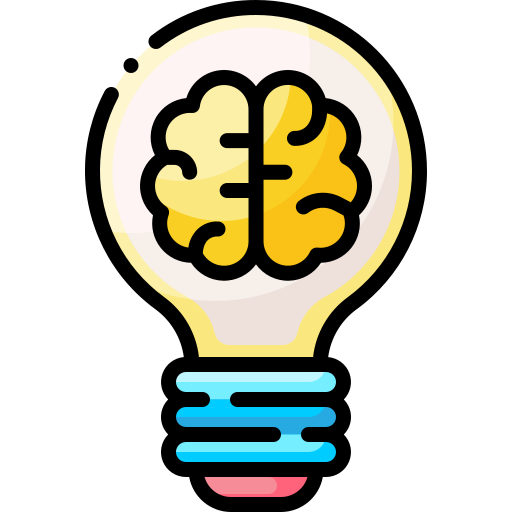}}

\newcommand{\framework}{\textsc{HintEval}\xspace}

\title{
    \begin{minipage}{2em}
        \centering
        \SpiritRAG
    \end{minipage}
    \hspace{-0.9em}
    \begin{minipage}{0.85\textwidth}
        \centering
        \raisebox{-1.7\height}{
        \shortstack{\framework: An Open-Source Python Toolkit for Hint Generation \\ and Hint Evaluation
        }
        }
    \end{minipage}
}

\newcommand{\JMorcid}{\orcidlink{0000-0003-4850-9239}}
\newcommand{\BPorcid}{\orcidlink{0009-0005-3578-2393}}
\newcommand{\AJorcid}{\orcidlink{0000-0001-7235-0665}}
\newcommand{\AAorcid}{\orcidlink{0000-0001-8747-4927}}

\author{Jamshid Mozafari\thanks{\, Corresponding Author.}\JMorcid, Bhawna Piryani \BPorcid, Abdelrahman Abdallah \AAorcid, Adam Jatowt \AJorcid \\
  University of Innsbruck, Innsbruck, Austria \\
  \texttt{\{jamshid.mozafari, adam.jatowt\}@uibk.ac.at} \\
}

\begin{document}
\maketitle

\begin{abstract}
Large Language Models (LLMs) increasingly provide direct answers to user questions, raising concerns about reduced engagement in critical thinking and problem-solving. Hint generation offers an alternative by guiding users toward answers without revealing them, while hint evaluation assesses the quality of such guidance.
Research in this area is hindered by fragmented datasets, inconsistent annotation formats, and evaluation tools that are often dataset-specific or unavailable. To address these challenges, we introduce \framework\footnote{\url{https://github.com/DataScienceUIBK/HintEval}}, an open-source Python library\footnote{\url{https://pypi.org/project/hinteval/}} for unified hint generation and evaluation. \framework standardizes access to diverse hint datasets, supports answer-aware and answer-agnostic generation methods, and implements multiple evaluation metrics within a shared data model. The toolkit enables reproducible experimentation, cross-dataset analysis, and multi-dimensional evaluation with minimal engineering effort. We further demonstrate its utility through human studies in which participants assess generated hints and use them to answer questions, showing that hints can effectively support users in reaching correct answers. \framework is accompanied by comprehensive documentation\footnote{\url{https://hinteval.readthedocs.io/}}, an executable Google Colab notebook for rapid experimentation, and a demonstration video\footnote{\url{https://youtu.be/QTPvW4V_6KY}}. By promoting consistent evaluation practices and lowering barriers to entry, it facilitates systematic research on hint-based question answering (QA) in NLP and IR.

\end{abstract}

\section{Introduction}\label{s:introduction}

\begin{figure*}[t]
\centering
\includegraphics[width=\textwidth]{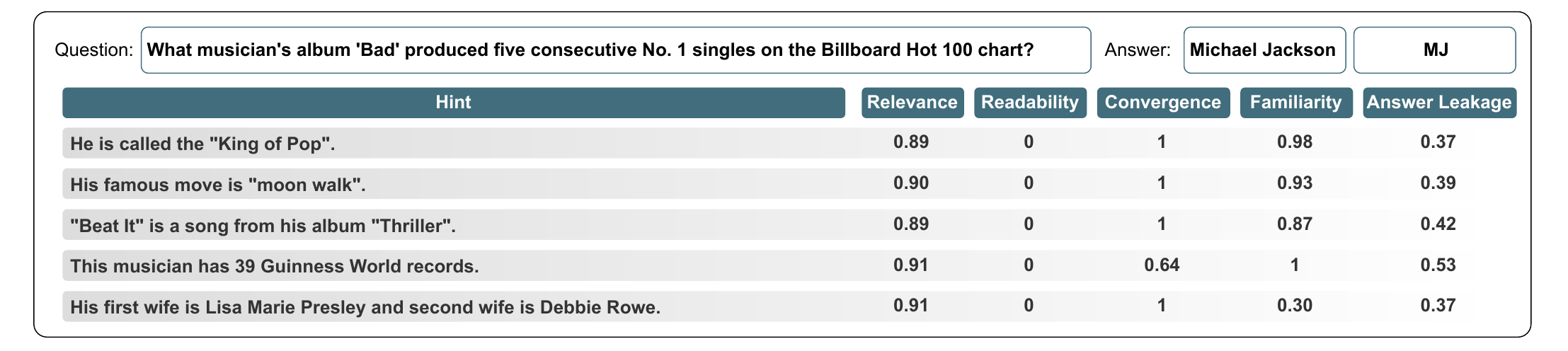}
\caption{Example hints for a sample question with scoring metrics. The metrics Relevance, Convergence, Familiarity, and Answer Leakage are rated on a scale from 0 to 1, where 0 represents the lowest and 1 the highest value. Higher scores in Relevance, Convergence, and Familiarity indicate better results, while a lower score is preferable for Answer Leakage. The Readability metric is scored as 0 (Beginner), 1 (Intermediate), or 2 (Advanced), with lower values indicating better readability.}
\label{fig:hint_sample}
\end{figure*}

Large Language Models (LLMs) have transformed Information Retrieval (IR) and Natural Language Processing (NLP) by enabling users to obtain answers to diverse questions quickly and conveniently~\cite{2023arXiv231211805G,2023arXiv230308774O,2024arXiv240721783D,2024arXiv240512819Q}. While this paradigm improves access to information, it has also raised concerns about reduced engagement in critical thinking, reasoning, and memory retention due to increasing reliance on AI systems~\cite{DARVISHI2024104967,Heersmink2024,10.1145/3706598.3713778}. This phenomenon, often referred to as \emph{cognitive offloading}, highlights the need for AI systems that support rather than replace human reasoning~\cite{singh2025protecting,soc15010006}.

To address this challenge, recent AI systems have begun exploring cognition-based interaction paradigms that guide users toward answers instead of providing them directly. Examples include ChatGPT \textit{Study and Learn}\footnote{\url{https://openai.com/index/chatgpt-study-mode/}} mode and Google \textit{LearnLM}\footnote{\url{https://cloud.google.com/solutions/learnlm}}. These approaches reflect growing interest in \emph{hint-based} question answering (QA), where hints provide clues that help users solve without explicitly revealing the answer~\cite{hume1996hinting}. Such interactions encourage active reasoning while retaining the benefits of AI assistance.

Research in this area focuses on two complementary tasks: \emph{Hint Generation} and \emph{Hint Evaluation}. Hint Generation aims to produce guidance that leads users toward the correct answer without disclosing it~\cite{2024arXiv240404728J}, while Hint Evaluation assesses hint quality along several dimensions~\cite{jatowt_kg_hint, bandura2013role}. Figure~\ref{fig:hint_sample} shows example hints and their evaluations.

Despite growing interest in hint-based methods~\cite{mozafari2026inferentialqa}, research progress remains hindered by fragmented datasets, heterogeneous annotation formats, and evaluation tools that are often dataset-specific or unavailable. Existing studies frequently introduce their own data formats, generation pipelines, and implementations, making it difficult to compare methods, reproduce results, and conduct cross-dataset analyses. As a result, \textit{comparing generation methods, studying evaluation metrics, and reproducing experimental findings require substantial engineering effort}.

To address these challenges, we introduce \framework, the first unified and extensible toolkit for hint generation and evaluation in NLP and IR. \framework standardizes access to diverse hint datasets, provides consistent interfaces for answer-aware and answer-agnostic hint generation, and implements widely used evaluation metrics within a shared data model. The toolkit enables reproducible experimentation, cross-dataset analysis, and large-scale evaluation with \textit{minimal engineering overhead}. We further demonstrate its usefulness through human-centered evaluations. It is released as open-source software with documentation, preprocessed datasets, a Google Colab notebook, and reproducible pipelines. The main contributions of this paper are:

\begin{enumerate}

\item We introduce \framework, the first unified toolkit that standardizes datasets, generation interfaces, and evaluation metrics for hint-based research.

\item We provide a unified implementation of diverse hint evaluation metrics for systematic hint assessment.

\item We demonstrate the toolkit through automatic and human-centered studies of hint quality and effectiveness.

\item We release \framework as open-source software with documentation, preprocessed datasets, a Google Colab notebook, and reproducible pipelines.

\end{enumerate}

\begin{figure*}[t]
  \centering
  \includegraphics[width=0.9\textwidth]{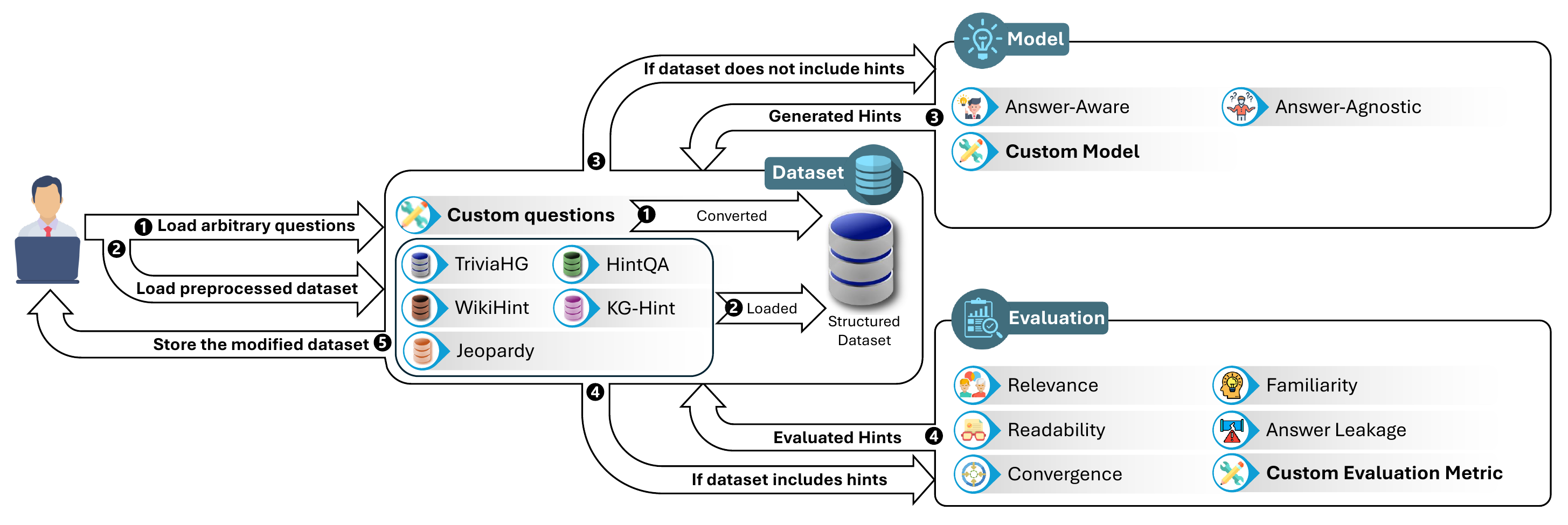}
  \caption{Workflow of \framework: \blackcircle{1}Questions are loaded and converted into a structured dataset using the \texttt{Dataset} module. \blackcircle{2}Users can load preprocessed datasets as a structured dataset.\blackcircle{3}Hints can be generated for each question using the \texttt{Model} module and stored in the dataset object. \blackcircle{4}The \texttt{Evaluation} module assesses all generated hints and questions using various evaluation metrics, storing the results in the dataset object. \blackcircle{5}The updated dataset can be saved and reloaded as needed.}
  \label{fig:hinteval_framework}
\end{figure*}

\section{Related Work}\label{s:related_work}

Hint generation originated in intelligent tutoring systems, where hints were designed to support reasoning and learning by guiding students toward solutions rather than providing direct answers~\cite{10.1145/3469885,Kochmar2022,Price2019}. More recently, hint-based interactions have been adapted to QA, with studies exploring answer-aware and answer-agnostic generation strategies, dataset construction, and automatic evaluation methods~\cite{mozafari-hintqa, mozafari-triviahg, 2024arXiv241201626M}. A comprehensive overview of recent advances in hint generation is provided 
by~\citet{2024arXiv240404728J}.

Most existing work builds on factual QA benchmarks such as TriviaQA~\cite{joshi-etal-2017-triviaqa}, Natural Questions~\cite{kwiatkowski-etal-2019-natural}, and WebQuestions~\cite{berant-etal-2013-semantic}. Representative resources, including TriviaHG~\cite{mozafari-triviahg}, HintQA~\cite{mozafari-hintqa}, WikiHint~\cite{2024arXiv241201626M}, KG-Hint~\cite{jatowt_kg_hint}, differ in their data sources, annotation schemes, and generation settings. Early approaches relied on human-authored clues or knowledge graphs, whereas recent work employs LLMs for scalable hint generation~\cite{2024arXiv240721783D,jangra2025designing}. Hint evaluation further extends traditional QA evaluation by considering dimensions such as Relevance, Readability, Convergence, Familiarity, and Answer Leakage~\cite{mozafari-triviahg,2024arXiv241201626M,mozafari2026context}. However, existing implementations are typically dataset-specific, making systematic comparison across datasets and generation methods difficult.

\begin{figure}[t!]
\centering
\includegraphics[width=\columnwidth]{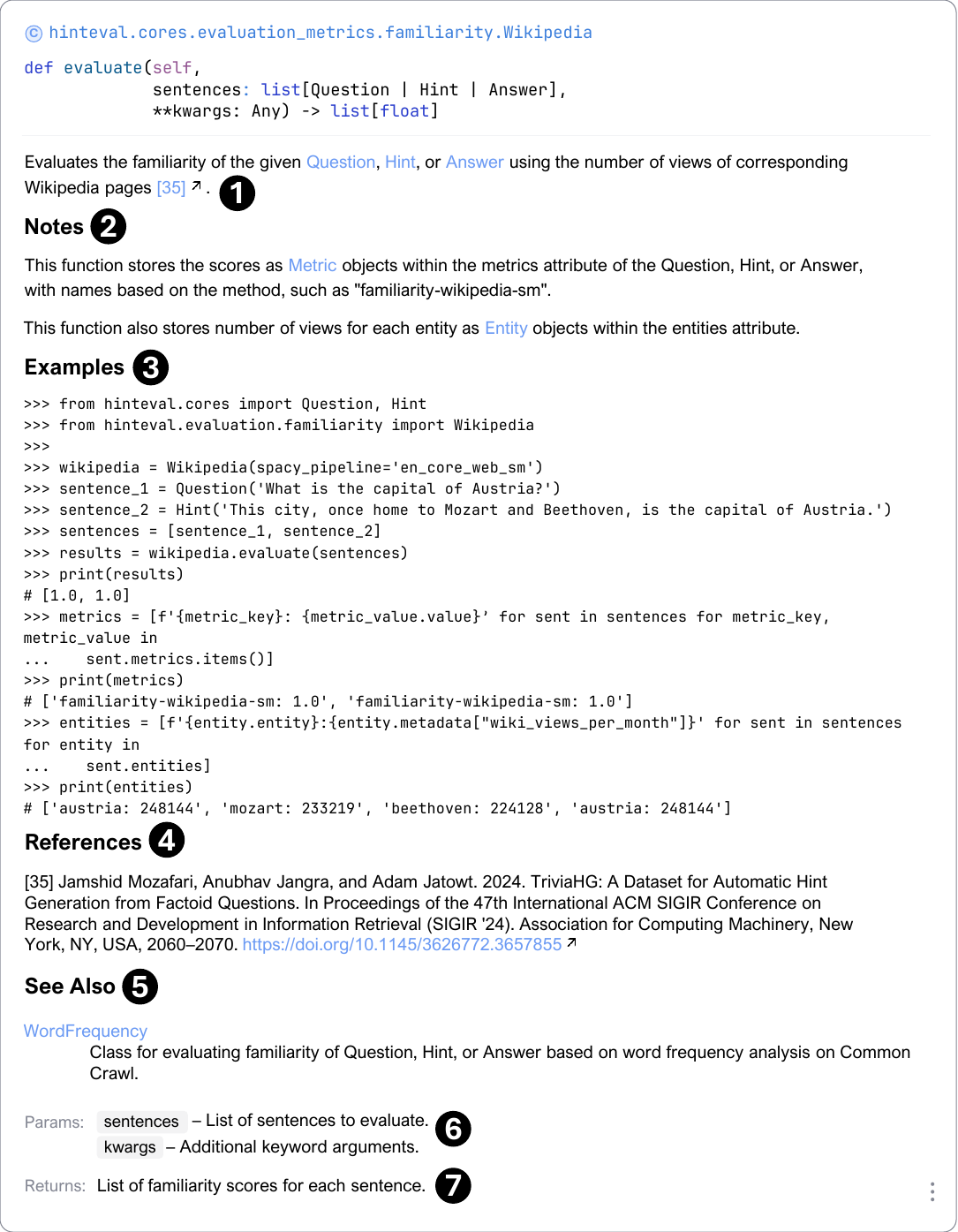}
\caption{Docstring for the \texttt{evaluate} function of the \texttt{Wikipedia} method within the \texttt{Familiarity} metric. It includes \blackcircle{1}a function description, \blackcircle{2}metric- and method-specific notes, and \blackcircle{3}a usage example. The documentation also provides \blackcircle{4}references to relevant publications, \blackcircle{5}links to related functions, \blackcircle{6}parameter descriptions, and \blackcircle{7}return value information.
}
\label{fig:docstring}
\end{figure}

General NLP frameworks support model development and experimentation, but they are not designed for hint-oriented workflows~\cite{10.1145/3404835.3463238,ushio-etal-2023-practical}. Existing hint-related work typically releases datasets, generation methods, or evaluation procedures independently. To the best of our knowledge, no prior toolkit unifies hint datasets, hint generation, and multi-dimensional hint evaluation within a single extensible framework. This gap motivates the development of \framework.

\section{\framework}\label{s:hint_eval}

\framework is a Python toolkit designed to simplify hint generation and evaluation, distributed via PyPI\footnote{\url{https://pypi.org/project/hinteval/}}, and designed to integrate naturally with common machine learning workflows based on libraries such as PyTorch~\cite{10.5555/3454287.3455008}. The toolkit can be installed using:
\begin{lstlisting}[numbers=none]
$ pip install hinteval
\end{lstlisting}

This paper introduces version \texttt{0.1.0}, the latest public release. \framework is open-source\footnote{released under the Apache-2.0 License.} and actively maintained. The toolkit provides documentation, preprocessed datasets, reproducible examples, and a Google Colab notebook demonstrating dataset loading, hint generation, and evaluation workflows. These resources enable users to reproduce the experiments presented in this paper with minimal setup effort.

\framework consists of three main modules: \textit{Dataset}, \textit{Model}, and \textit{Evaluation}. The \textit{Dataset} module standardizes access to heterogeneous hint datasets, the \textit{Model} module supports hint generation, and the \textit{Evaluation} module provides a unified interface for computing hint-related metrics. Figure~\ref{fig:hinteval_framework} illustrates the overall workflow.

All public classes and methods in \framework are documented using structured docstrings that are automatically rendered in the online documentation. Figure~\ref{fig:docstring} illustrates an example documentation page for the Wikipedia-based familiarity evaluator, including descriptions, usage examples, references, cross-links, and parameter and return specifications.

\input{tables/table_1}

\subsection{Dataset}\label{ss:datasets}

\begin{figure}[t]
  \centering
  \includegraphics[width=\columnwidth]{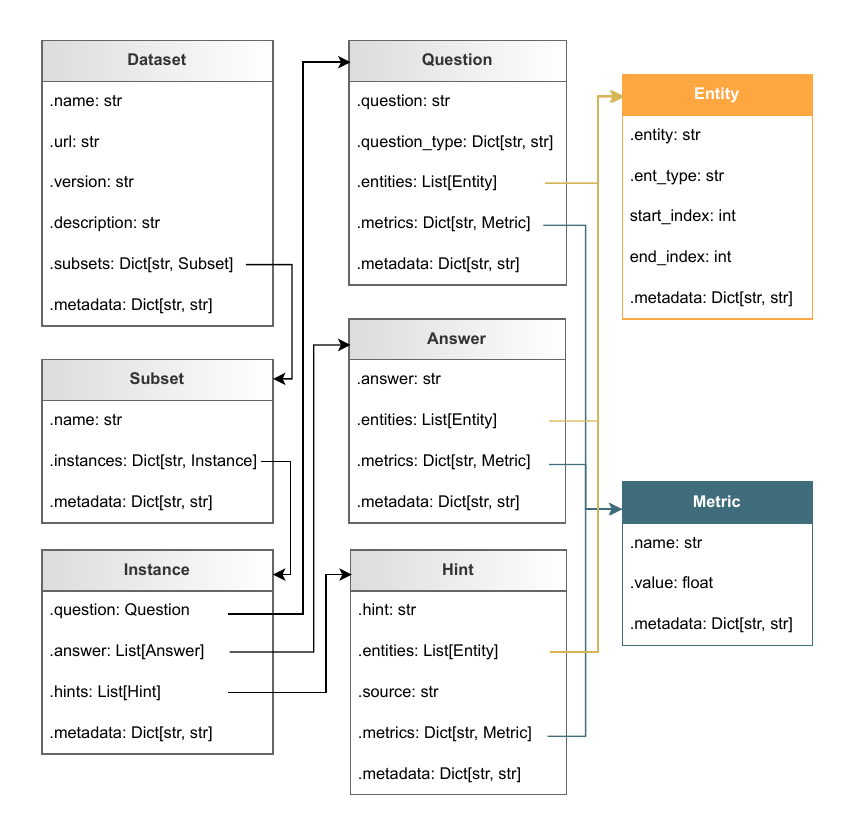}
  \caption{Schema of the \texttt{Dataset} class, illustrating the objects used to represent hint datasets in \framework. Arrows indicate relationships between the objects.}
  \label{fig:dataset_schema}
\end{figure}

A central goal of \framework is to enable seamless experimentation across multiple hint datasets with minimal engineering effort. To this end, we introduce a unified dataset abstraction that harmonizes diverse annotation styles and storage formats, as
shown in Figure~\ref{fig:dataset_schema}.

Each dataset comprises one or more subsets (e.g., train, validation, test), where each subset contains a collection of instances indexed by unique question IDs. An instance consists of a \texttt{Question}, one or more \texttt{Answer} objects, and a set of \texttt{Hint} objects. This structure supports questions with multiple valid answers and multiple associated hints.

The \texttt{Question}, \texttt{Answer}, and \texttt{Hint} classes share common attributes for entities and evaluation metrics. Named entities are automatically extracted using spaCy~\cite{Honnibal_spaCy_Industrial-strength_Natural_2020}, while metric values are stored in a unified format to support consistent analysis across datasets and evaluation methods. Dataset-specific features are preserved through a flexible \texttt{metadata} field.

\framework provides preprocessed versions of several existing hint datasets, which can be downloaded and loaded with a few lines of code. Users may also define custom datasets using the same abstraction. Datasets can be saved and loaded using JSON format, enabling easy reuse and sharing across experiments. Table~\ref{tbl:preproccesed_datasets} summarizes the preprocessed datasets currently supported by \framework. The supported datasets cover human-authored, knowledge-graph-based, and LLM-generated hints, as well as both answer-aware and answer-agnostic generation settings. 

To demonstrate the dataset construction capabilities of \framework, we provide a large-scale Jeopardy-based resource containing 68,886 questions and 347,484 hints. The resource is derived from original Jeopardy clues, which are treated as hints, while questions are generated using LLMs and subsequently evaluated using \framework. This example demonstrates how \framework simplifies the construction and evaluation of new hint datasets while supporting large-scale experimentation. Additional details of the dataset creation pipeline are provided in Appendix~\ref{apx:jeopardy}.

\begin{figure*}
  \centering
  \includegraphics[width=0.9\textwidth]{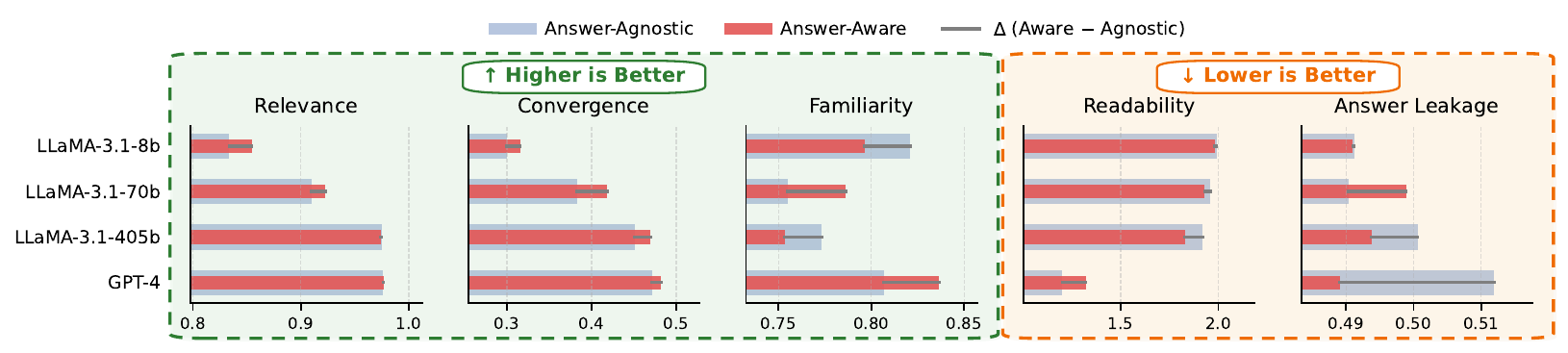}
  \caption{Comparison of hint generation strategies across four LLMs using evaluation metrics.}
  \label{fig:aware-vs-agnostic}
\end{figure*}

\subsection{Model}\label{ss:models}

\framework includes two built-in hint generation models: \emph{Answer-Aware} and \emph{Answer-Agnostic}.

The \emph{Answer-Aware} model generates hints given both a question and its ground-truth answer. While this setting typically produces higher-quality hints~\cite{2024arXiv241201626M}, it assumes that gold answers are available. This approach was used to construct datasets such as TriviaHG~\cite{mozafari-triviahg}. In \framework, this model is implemented via the \texttt{AnswerAware} class, which populates instances with generated hints. In practice, \textit{it can support teachers in preparing lesson materials}.

The \emph{Answer-Agnostic} model generates hints using only the question, making it applicable when answers are unavailable. Although more flexible, this approach may yield lower-quality hints due to uncertainty about the correct answer~\cite{2024arXiv241201626M}. Datasets such as HintQA~\cite{mozafari-hintqa} adopt this setting, while WikiHint~\cite{2024arXiv241201626M} combines both approaches. The model is implemented as the \texttt{AnswerAgnostic} class. In practice, \textit{it can help users avoid cognitive offloading by providing hints instead of direct answers}.

Both models inherit from a \texttt{Model} base class and expose a unified \texttt{generate} interface, enabling integration with datasets and evaluation modules. 
Users can extend the toolkit by implementing custom models that conform to the \texttt{Model} interface.

\subsection{Evaluation}\label{ss:evaluation}

Hint evaluation assesses the quality of generated hints using multiple complementary criteria. \framework implements five evaluation dimensions--\emph{Relevance}, \emph{Readability}, \emph{Convergence}, \emph{Familiarity}, and \emph{Answer Leakage}--through fifteen evaluation methods and thirty-five sub-methods. \emph{Relevance} measures the semantic relatedness between a hint and its question, \emph{Readability} captures the ease with which a hint can be understood, \emph{Convergence} evaluates how effectively a hint narrows the space of possible answers, \emph{Familiarity} estimates how common or well-known the hinted information is, and \emph{Answer Leakage} assesses the extent to which a hint reveals the answer.

Early work introduced evaluation metrics for Convergence and Familiarity~\cite{mozafari-triviahg}, while subsequent studies proposed methods for Relevance, Readability, and Answer Leakage~\cite{2024arXiv241201626M}. However, existing implementations are dataset-specific, computationally expensive, or tightly coupled to particular schemas.

\framework implements all evaluation metrics in a dataset-agnostic manner, enabling them to operate on any dataset represented using the \texttt{Dataset} abstraction. In addition, \framework introduces multiple evaluation methods per metric, ranging from lightweight statistical approaches to neural and LLM-based methods, allowing users to balance accuracy and computational cost.
Table~\ref{tbl:metrics_comparison} in the Appendix compares the supported evaluation methods across metrics. The assessments are qualitative guidelines based on the computational characteristics reported in the original publications and our experience implementing and using the methods within \framework. They are intended to assist users in selecting evaluation methods rather than to serve as formal benchmark measurements.

Evaluation metrics in \framework inherit from a common \texttt{Evaluation} base class. Users can define custom metrics by implementing the \texttt{evaluate} function, enabling seamless extension and integration with the existing evaluation pipeline.

\begin{figure}[t]
\begin{lstlisting}[
    numbers=none,
    aboveskip=0pt,
    belowskip=0pt,
    caption={An example showing the \framework workflow for creating a custom dataset (\texttt{own\_dataset}) with two instances and a \texttt{test} subset, generating three hints using \texttt{AnswerAware} with Qwen3-8B, evaluating them with five metrics, and saving the results to \texttt{dataset.json}.},
    label={lst:evaluator},
    language=Python
]
from hinteval import Dataset
from hinteval.cores.dataset_core import Instance, Subset
from hinteval.model import AnswerAware, AnswerAgnostic
from hinteval.evaluation.relevance import Rouge
from hinteval.evaluation.readability import MachineLearningBased
from hinteval.evaluation.convergence import NeuralNetworkBased
from hinteval.evaluation.familiarity import Wikipedia
from hinteval.evaluation.answer_leakage import Lexical

""""""" Dataset """""""
instance_1 = Instance.from_strings('Q1', ['A11'], [])
instance_2 = Instance.from_strings('Q2', ['A21'], [])

subset = Subset('test')
subset.add_instance(instance_1)
subset.add_instance(instance_2)

dataset = Dataset('own_dataset')
dataset.add_subset(subset)

""""""" Model """""""
instances = dataset['test'].get_instances()

answer_aware = AnswerAware('Qwen/Qwen3-8B', num_of_hints=3)
answer_aware.generate(instances)

""""""" Evaluation """""""
relevance = Rouge('rougeL')
    .evaluate(instances)
readability = MachineLearningBased('xgboost')
    .evaluate(instance_1.hints)
convergence = NeuralNetworkBased('bert-base')
    .evaluate(instances)
familiarity = Wikipedia()
    .evaluate(instance_2.hints)
answer_leakage = Lexical('exclude_stop_words')
    .evaluate(instances)

""""""" Store """""""
dataset.store_json('./path/to/dataset.json')
\end{lstlisting}
\vspace{-1em}
\end{figure}

Listing~\ref{lst:evaluator} presents a code example using \framework that demonstrates the three main modules of the pipeline, from dataset construction to result storage. \framework supports running LLMs either locally or via remote APIs, allowing users to select an execution mode based on their available computational resources. Models hosted on HuggingFace\footnote{\url{https://huggingface.co/}} can be used seamlessly. For further details, we refer readers to the online documentation, which provides comprehensive explanations and illustrative examples.

\section{Experiments and Results}\label{s:experiments}

We demonstrate three representative capabilities of \framework. Specifically, we show how the toolkit supports systematic comparison of hint generation strategies, exploratory analyses of evaluation metrics, and validation of generated hints through human studies. These examples illustrate typical research workflows enabled by \framework.

\subsection{Comparing Hint Generation Strategies}\label{ss:strategies_comparison}

\framework provides a unified evaluation pipeline that enables comparison of hint generation strategies across models and datasets. As an example, we compare Answer-Aware and Answer-Agnostic generation using four LLMs on the WikiHint dataset.

Figure~\ref{fig:aware-vs-agnostic} shows that Answer-Aware generation generally improves Relevance, Convergence, and Readability while reducing Answer Leakage. The differences become smaller for stronger LLMs, whereas Familiarity remains largely unchanged.

This example\footnote{\url{https://github.com/DataScienceUIBK/HintEval/blob/main/experiments/aware-vs-agnostic/aware-vs-agnostic.py}} demonstrates how \framework enables comparison of hint generation strategies using multiple complementary evaluation metrics. Researchers can readily apply the same pipeline to compare their own methods with existing approaches with minimal additional code.

\begin{figure*}
  \centering
  \includegraphics[width=\textwidth]{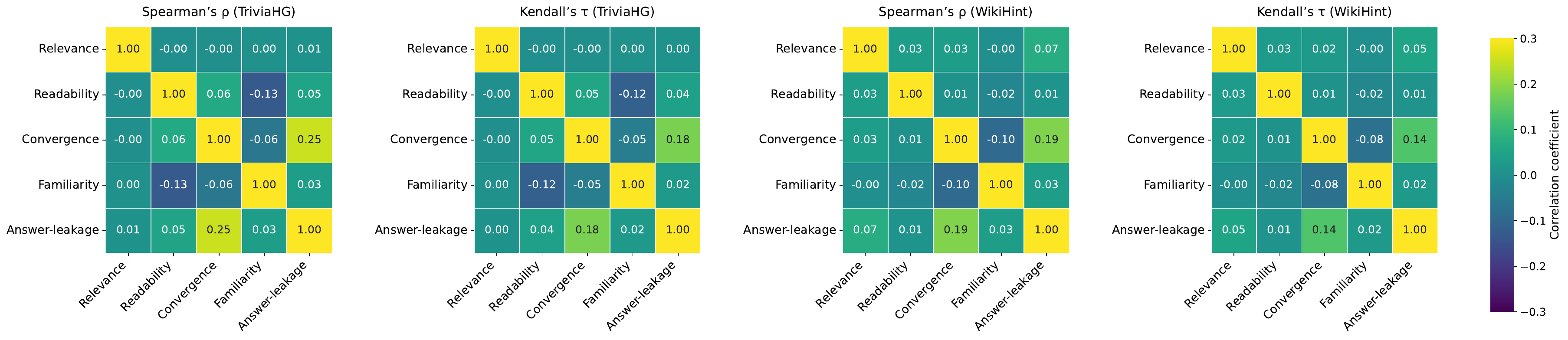}
  \caption{Correlation among evaluation metrics computed using Spearman's $\rho$ and Kendall's $\tau$ for the TriviaHG and WikiHint datasets.}
  \label{fig:metric-agreement}
\end{figure*}

\subsection{Analyzing Evaluation Metrics}\label{ss:metrics_comparison}

Beyond evaluating individual hints, \framework supports large-scale analyses of relationships among evaluation metrics.

Figure~\ref{fig:metric-agreement} presents correlations between the five evaluation dimensions computed on TriviaHG and WikiHint. Most correlations are weak and dataset-dependent, indicating that the metrics capture complementary aspects of hint quality. The only consistent trend is a positive correlation between Convergence and Answer Leakage, highlighting the inherent trade-off between producing informative hints and avoiding answer disclosure.

This example\footnote{\url{https://github.com/DataScienceUIBK/HintEval/blob/main/experiments/metric-agreement/metric-agreement.py}} illustrates how \framework enables researchers to easily analyze relationships among evaluation metrics on their own datasets using a unified analysis pipeline with minimal additional code.

\subsection{Human Evaluation}\label{ss:human_evaluation}

Automatic metrics provide scalable evaluation, but human assessment remains essential for validating the quality and usefulness of hints generated by \framework. We conducted two complementary studies focused on hint quality and effectiveness: \textbf{(1)} a hint quality assessment and \textbf{(2)} a hint-assisted question answering study. We additionally conducted a separate toolkit usability study to assess the ease of use of \framework. Details of the human evaluation protocols are provided in Appendix~\ref{apx:human_evaluation}.

\input{tables/table_4}
\input{tables/table_6}

\paragraph{Hint Quality Assessment.}

We randomly selected 500 questions from the TriviaQA~\cite{joshi-etal-2017-triviaqa} and NQ~\cite{kwiatkowski-etal-2019-natural} test sets covering diverse categories. For each question, we generated five hints using LLaMA-3.3-70B, resulting in 2,500 hints. To reduce answer leakage, we removed hints with an Answer Leakage score higher than 0.8, leaving 2,214 hints for evaluation.


Twenty human annotators participated in the study. The hints were distributed among annotators and assessed using a five-point Likert scale (1: very low, 5: very high) across the five dimensions implemented in \framework. Higher values indicate better quality for Relevance, Readability, Familiarity, and Convergence, whereas lower values indicate less Answer Leakage and are therefore preferable.

Table~\ref{tab:human_eval} reports the average ratings. Overall, annotators judged the generated hints to be relevant, understandable, and effective in guiding users toward the correct answer while avoiding direct answer disclosure. In particular, Relevance and Readability achieved average scores above 4, whereas Answer Leakage received a low score.

To assess whether the automatic evaluation metrics align with human judgment, we computed Spearman's $\rho$ between automatic metric scores and human ratings. As shown in Table~\ref{tbl:metric_human_corr}, all metrics exhibit moderate positive correlations (0.41--0.61), providing evidence that the evaluation methods implemented in \framework capture aspects of hint quality that are consistent with human assessments.

\input{tables/table_7}

\paragraph{Hint-Assisted Question Answering.}

To evaluate the usefulness of the generated hints, we recruited a separate group of thirty participants who had not taken part in the previous study.

Participants first attempted to answer the selected questions without hints. Questions answered correctly were excluded from the second phase, as no assistance was required. The remaining unanswered questions were then presented together with their corresponding hints.

Without hints, participants correctly answered 89 of the 500 questions (17.8\%). After receiving hints, they correctly answered 312 of the remaining 401 unanswered questions, corresponding to a success rate of 77.8\%. Overall question-answering accuracy increased from 17.8\% to 80.2\%.

These findings demonstrate that hints generated and evaluated using \framework effectively support users in answering questions while encouraging reasoning rather than directly revealing the answer.

\paragraph{Toolkit Usability Study.}

We conducted a usability study with ten students and researchers familiar with NLP or IR. Participants completed representative tasks using \framework, including dataset loading, hint generation, evaluation, and result export, before rating the toolkit on six five-point Likert-scale questions as shown in Table~\ref{tbl:usability}.

Overall, participants found \framework easy to learn and use, with positive ratings for the API, documentation, and example code, suggesting that the toolkit can be readily adopted for hint-generation research.

\section{Conclusion}\label{s:conclusion}

We presented \framework, an open-source toolkit for hint generation and evaluation that unifies datasets, generation interfaces, and evaluation metrics within a common framework. By standardizing data representations and evaluation workflows, \framework enables reproducible and cross-dataset experimentation with minimal engineering effort. Through automatic and human-centered studies, we demonstrated its utility for evaluating hint quality and supporting question answering.
Future work will extend the framework with additional generation methods and evaluation metrics, and support for emerging LLMs.

\section*{Ethical Considerations}

\framework is designed to support reasoning-focused interactions by providing hints rather than direct answers. As a research toolkit, it enables the generation and evaluation of hints for question-answering tasks, educational applications, and studies of human--AI interaction.

Several ethical considerations remain relevant. First, generated hints may inherit factual inaccuracies, hallucinations, or social biases from the underlying language models and datasets. Second, hints may inadvertently reveal answers, reducing their pedagogical value or enabling misuse in assessment settings. Third, the quality and fairness of generated hints may vary across domains, topics, and user populations.

To mitigate these risks, \framework incorporates multiple evaluation dimensions, including relevance, convergence, familiarity, readability, and answer leakage, allowing researchers and practitioners to assess hint quality and identify potentially problematic outputs. The toolkit is intended for research and educational purposes, and human oversight remains important when deploying hint-based systems in real-world settings.

\section*{Limitations}

While \framework provides a unified and extensible framework for hint generation and evaluation, several limitations remain in the current version and experimental evaluation.

\begin{itemize}
    \item \textbf{Domain coverage.} The current experiments primarily focus on factual and open-domain QA datasets. Further evaluation is needed to assess the generalizability of \framework to other domains and task settings.

    \item \textbf{Human evaluation.} The human studies involve relatively small groups of students and researchers, and the hint quality assessment is based on hints generated by a single LLM. These factors may limit the generalizability of the findings across broader user populations and generation models.
\end{itemize}

\section*{Acknowledgments} \label{s:acknowledgments}  
The computational results presented here have been achieved (in part) using the LEO HPC infrastructure of the University of Innsbruck and the Austrian Scientific Computing (ASC) infrastructure.


\newpage
\bibliography{custom}

\clearpage

\appendix

\input{tables/table_3}

\section{Jeopardy Dataset Construction}\label{apx:jeopardy}

To demonstrate the dataset construction capabilities of \framework, we created a large-scale Jeopardy-based resource, named \texttt{Jeopardy}, containing 68,886 questions and 347,484 hints.

The resource was derived from the publicly available Jeopardy clue collection\footnote{\url{https://github.com/jwolle1/jeopardy_clue_dataset}}. Since Jeopardy clues provide information that guides participants toward a correct response, we treat them as hints. For each clue, an LLM was prompted to generate a question. The resulting question--hint pairs were subsequently processed using \framework.

The construction pipeline consisted of four stages: \textbf{(1)} Collection of Jeopardy clues and associated metadata.
\textbf{(2)} Conversion of clues into hint instances using the \framework data model.
\textbf{(3)} Generation of questions using LLaMA 3.3 70B.
\textbf{(4)} Evaluation of the resulting hint--question pairs using \framework, including entity extraction and computation of all supported evaluation
metrics.

The final resource contains 68,886 questions, 347,484 hints, automatically extracted entities, and precomputed evaluation scores. The resulting dataset demonstrates how \framework can be used not only to evaluate existing hint datasets but also to construct, enrich, and analyze new hint-oriented resources at scale.

\section{Human Evaluation Protocol}\label{apx:human_evaluation}

\paragraph{Hint Quality Assessment.}

We randomly selected 500 questions from the TriviaQA and NQ test sets and generated five hints per question using LLaMA-3.3-70B. After filtering hints with an Answer Leakage score higher than 0.8, 2,214 hints remained.

Twenty annotators, consisting of students and researchers, participated in the study. To balance annotation quality and effort, the generated hints were distributed among annotators such that each hint was independently evaluated by two annotators. 
Annotators assessed each hint using a five-point Likert scale (1: very low, 5: very high) for Relevance, Readability, Familiarity, Convergence, and Answer Leakage. Higher ratings are preferable for Relevance, Readability, Familiarity, and Convergence, whereas lower ratings indicate less Answer Leakage and are therefore preferable.

Annotators were instructed to evaluate each hint independently and assign ratings according to their subjective judgment of each dimension.

\paragraph{Hint-Assisted Question Answering.}

A separate group of thirty participants, consisting of students and researchers, took part in the hint-assisted question answering study. Questions were distributed among participants.

The study followed a three-stage process:
\textbf{(1)} Participants first attempted to answer each question without using hints. If the answer was correct, they proceeded to the next question.
\textbf{(2)} If the answer was incorrect or unknown, participants were shown the corresponding hints and could review them sequentially while attempting to solve the question.
\textbf{(3)} If participants were still unable to answer after reviewing all available hints, they were allowed to skip the question.

The primary evaluation measure was QA accuracy before and after exposure to hints.

\section{Evaluation Methods and Sub-methods}
\label{apx:evaluation_methods}

\framework organizes hint evaluation into five dimensions, fifteen methods, and thirty-five sub-methods. The methods provide alternative ways to estimate each evaluation dimension, while sub-methods differ in model architecture, representation, provider, or preprocessing choice. Table~\ref{tbl:evaluation_methods} summarizes the complete mapping.

\input{tables/table_5}

Because these approaches rely on different signals, methods targeting the same dimension may produce different scores and should be interpreted as complementary rather than interchangeable.

\paragraph{Relevance.}
Relevance measures the semantic relationship between a hint and its corresponding question. \framework supports four method families. ROUGE-1, ROUGE-2, and ROUGE-L capture lexical overlap at the unigram, bigram, and sequence levels, respectively. The non-contextual method instead uses fixed GloVe representations, with GloVe-6B and GloVe-42B providing two alternative embedding resources. Contextual methods use BERT-base and RoBERTa-large to capture semantic relationships beyond direct word overlap. In the reported evaluation, RoBERTa-large achieved higher MAP and MRR than BERT-base, while GloVe-42B slightly outperformed GloVe-6B. LLM-based variants provide another formulation in which models accessed through HuggingFace, TogetherAI, or OpenAI are used to assess question--hint relevance.

\paragraph{Readability.}
Readability measures how easily a hint can be understood. Traditional sub-methods include Gunning-Fog, Flesch Reading Ease, Coleman-Liau, SMOG, and Automated Readability Index, all of which estimate difficulty from surface-level textual characteristics. Machine-learning alternatives use XGBoost and Random Forest, while neural approaches use BERT-base and RoBERTa-large to predict readability from contextualized text representations. In the reported experiments, XGBoost performed slightly better than Random Forest, and RoBERTa-large achieved higher accuracy and F1 than BERT-base. LLM-based methods provide an additional option for directly judging the readability of a hint.

\paragraph{Convergence.}
Convergence captures how effectively a hint narrows the space of plausible answers. The specificity-based method uses BERT-base and RoBERTa-large to estimate how specific or general the information in a hint is. Neural convergence methods use the same model families but are trained directly on convergence values, making them conceptually different from the specificity predictors despite sharing similar architectures. In the reported evaluation, their predicted convergence values achieved Pearson correlations of 0.561 for BERT-base and 0.611 for RoBERTa-large with the ground-truth values. The LLM-based approach follows a different strategy by generating candidate answers and evaluating how strongly a hint reduces the set of plausible candidates.

\paragraph{Familiarity.}
Familiarity estimates how common or well-known the information expressed in a hint is. The word-frequency method uses corpus-based lexical frequency and is implemented both with and without stop words. The two variants share the same underlying formulation but differ in whether frequent functional words contribute to the resulting score. The Wikipedia-based method instead focuses on entities and uses the popularity of their corresponding Wikipedia pages as an indicator of familiarity. These approaches therefore capture different aspects of familiarity: lexical commonness and entity-level popularity.

\paragraph{Answer Leakage.}
Answer Leakage measures the extent to which a hint reveals its corresponding answer. The lexical method captures explicit word-level overlap between hints and answers and provides variants with and without stop words. The contextual method instead uses semantic representations, allowing it to detect similarity even when the hint and answer use different surface forms. Consequently, lexical and contextual approaches provide complementary views of answer disclosure, with the former emphasizing direct overlap and the latter capturing semantically related expressions.

\paragraph{Relationship among sub-methods.}
Sub-methods within the same method typically differ in a narrow modeling or preprocessing choice, such as transformer architecture, embedding resource, model size, provider, or stop-word handling. In contrast, different methods within the same evaluation dimension may rely on different signals. For example, Relevance can be estimated through lexical overlap, static embeddings, contextual representations, or LLM judgments, while Familiarity can be estimated through word frequency or Wikipedia-based entity popularity.

The correlation analysis in Section~\ref{ss:metrics_comparison} focuses on relationships among the five high-level evaluation dimensions using Spearman's $\rho$ and Kendall's $\tau$, rather than agreement among individual sub-methods. 

\end{document}

%% file: tables/table_1.tex
\begin{table}[t]
\centering
\small
\resizebox{\columnwidth}{!}{%
\begin{tabular}{lrrlcc}
\toprule
Dataset & Questions & Hints & Source & Aw. & Ag. \\
\midrule
TriviaHG
& 16,645
& 160,228
& LLM
& \ding{51}
& \ding{55} \\

WikiHint
& 1,000
& 5,000
& Human/LLM
& \ding{51}
& \ding{51} \\

HintQA
& 16,955
& 305,624
& LLM
& \ding{55}
& \ding{51} \\

KG-Hint
& 30
& 307
& KG
& \ding{55}
& \ding{51} \\

Jeopardy
& 68,886
& 347,484
& Human/Machine
& \ding{55}
& \ding{51} \\
\bottomrule
\end{tabular}
}%
\caption{Summary of the preprocessed hint datasets supported by \framework. Columns \textit{Aw.} and \textit{Ag.} indicate whether the dataset contains answer-aware and answer-agnostic hints, respectively.}
\label{tbl:preproccesed_datasets}
\end{table}

%% file: tables/table_4.tex
\begin{table}[t]
\centering
\resizebox{\columnwidth}{!}{%
\begin{tabular}{lccccc}
\toprule
Metric & Relevance & Readability & Familiarity & Convergence & Answer Leakage \\
\midrule
Mean Rating & 4.01 & 4.47 & 3.81 & 3.57 & 1.76 \\
\bottomrule
\end{tabular}
}
\caption{Human evaluation results on 2,214 hints. Ratings are on a five-point Likert scale (1: very low, 5: very high). Higher scores are preferable for Relevance, Readability, Familiarity, and Convergence, whereas lower scores are preferable for Answer Leakage.}
\label{tab:human_eval}
\end{table}

%% file: tables/table_6.tex
\begin{table}[t]
\centering
\resizebox{\columnwidth}{!}{%
\begin{tabular}{lccccc}
\toprule
Metric & Relevance & Readability & Familiarity & Convergence & Answer Leakage \\
\midrule
Spearman's $\rho$ & 0.45 & 0.41 & 0.53 & 0.61 & 0.53 \\
\bottomrule
\end{tabular}
}
\caption{Correlation between automatic evaluation metrics and human ratings. Spearman's $\rho$ is computed using the subset of hints evaluated by human annotators.}
\label{tbl:metric_human_corr}
\end{table}

%% file: tables/table_7.tex
\begin{table}[t]
\centering
\resizebox{\columnwidth}{!}{
\begin{tabular}{l@{\hspace{100pt}}c}
\toprule
\textbf{Statement} & \textbf{Mean} \\
\midrule
The toolkit was easy to install. & 4.8 \\
The API was easy to understand. & 4.2 \\
The documentation was helpful. & 4.7 \\
The provided examples were useful. & 4.5 \\
The toolkit was easy to use. & 4.2 \\
I would use \framework\ in future research. & 3.8 \\
\bottomrule
\end{tabular}}
\caption{Usability evaluation of \framework.}
\label{tbl:usability}
\end{table}

%% file: tables/table_3.tex
\begin{table}[t]
\resizebox{\columnwidth}{!}{%
\begin{tabular}{@{}llllll@{}}
\toprule
Metric                         & Method           & \multicolumn{1}{c}{\begin{tabular}[c]{@{}c@{}}Preferred\\ Device\end{tabular}} & \multicolumn{1}{c}{\begin{tabular}[c]{@{}c@{}}Cost\\ Effectiveness\end{tabular}} & Accuracy  & \multicolumn{1}{c}{\begin{tabular}[c]{@{}c@{}}Execution\\ Speed\end{tabular}} \\ \midrule
\multirow{4}{*}{Relevance}     & ROUGE            & CPU                                                                            & High                                                                             & Low       & Very Fast                                                                     \\
                               & Non-Contextual   & CPU/GPU                                                                        & High                                                                             & Moderate  & Fast                                                                          \\
                               & Contextual       & GPU                                                                            & Moderate                                                                         & High      & Moderate                                                                      \\
                               & LLM              & GPU                                                                            & Low                                                                              & Very High & Slow                                                                          \\ \midrule
\multirow{4}{*}{Readability}   & Traditional      & CPU                                                                            & High                                                                             & Low       & Very Fast                                                                     \\
                               & Machine-Learning & CPU                                                                            & High                                                                             & Moderate  & Moderate                                                                      \\
                               & Neural-Network   & GPU                                                                            & Moderate                                                                         & High      & Moderate                                                                      \\
                               & LLM              & GPU                                                                            & Low                                                                              & Very High & Slow                                                                          \\ \midrule
\multirow{3}{*}{Convergence}   & Specificity      & GPU                                                                            & Moderate                                                                         & Low       & Moderate                                                                      \\
                               & Neural-Network   & GPU                                                                            & Moderate                                                                         & Moderate  & Moderate                                                                      \\
                               & LLM              & GPU                                                                            & Low                                                                              & High      & Slow                                                                          \\ \midrule
\multirow{2}{*}{Familiarity}   & Word-Frequency   & CPU                                                                            & Very High                                                                        & Low       & Very Fast                                                                     \\
                               & Wikipedia        & CPU                                                                            & High                                                                             & High      & Slow                                                                          \\ \midrule
\multirow{2}{*}{Answer Leakage} & Lexical          & CPU                                                                            & Very High                                                                        & Low       & Very Fast                                                                     \\
                               & Contextual       & GPU                                                                            & Moderate                                                                         & High      & Moderate                                                                      \\ \bottomrule
\end{tabular}%
}
\caption{Comparison of evaluation methods implemented in \framework. Accuracy, cost-effectiveness, and speed are qualitative assessments intended to guide method selection.}
\label{tbl:metrics_comparison}
\end{table}

%% file: tables/table_5.tex
\begin{table}[t]
\centering
\small
\resizebox{\columnwidth}{!}{%
\begin{tabular}{lp{9cm}}
\toprule
\textbf{Metric} & \textbf{Implemented Methods} \\
\midrule

Relevance &
ROUGE-1, ROUGE-2, ROUGE-L, GloVe-6B, GloVe-42B, BERT-base, RoBERTa-large, HuggingFace LLMs, TogetherAI LLMs, OpenAI GPT models \\
\midrule

Readability &
Gunning-Fog Index, Flesch Reading Ease, Coleman-Liau Index, SMOG Index, Automated Readability Index, XGBoost, Random Forest, BERT-base, RoBERTa-large, HuggingFace LLMs, TogetherAI LLMs, OpenAI GPT models \\
\midrule

Convergence &
Specificity (BERT-base), Specificity (RoBERTa-large), BERT-base, RoBERTa-large, LLaMA 3.1 8B, LLaMA 3.1 70B \\
\midrule

Familiarity &
Word Frequency (with stop words), Word Frequency (without stop words), Wikipedia-based Familiarity \\
\midrule

Answer Leakage &
Lexical Overlap (with stop words), Lexical Overlap (without stop words), Contextual Similarity (with stop words), Contextual Similarity (without stop words) \\
\bottomrule
\end{tabular}
}%
\caption{Evaluation methods currently implemented in \framework, grouped by evaluation metric.}
\label{tbl:evaluation_methods}
\end{table}

%% file: custom.bib
@ARTICLE{2023arXiv231211805G,
       author = {{Gemini Team} and {Anil}, Rohan and {Borgeaud}, Sebastian and {Alayrac}, Jean-Baptiste and {Yu}, Jiahui and {Soricut}, Radu and {Schalkwyk}, Johan and {Dai}, Andrew M. and {Hauth}, Anja and {Millican}, Katie and {Silver}, David and {Johnson}, Melvin and Others},
        title = "{Gemini: A Family of Highly Capable Multimodal Models}",
      journal = {arXiv e-prints},
         year = 2023,
        month = dec,
          eid = {arXiv:2312.11805},
        pages = {arXiv:2312.11805},
          doi = {10.48550/arXiv.2312.11805}
}

@ARTICLE{2023arXiv230308774O,
       author = {{OpenAI} and {Achiam}, Josh and {Adler}, Steven and {Agarwal}, Sandhini and {Ahmad}, Lama and {Akkaya}, Ilge and {Leoni Aleman}, Florencia and {Almeida}, Diogo and {Altenschmidt}, Janko and {Altman}, Sam and Others},
        title = "{GPT-4 Technical Report}",
      journal = {arXiv e-prints},
         year = 2023,
        month = mar,
          eid = {arXiv:2303.08774},
        pages = {arXiv:2303.08774},
          doi = {10.48550/arXiv.2303.08774}
}

@article{2024arXiv240721783D,
       author = {{Dubey}, Abhimanyu and {Jauhri}, Abhinav and {Pandey}, Abhinav and {Kadian}, Abhishek and {Al-Dahle}, Ahmad and {Letman}, Aiesha and {Mathur}, Akhil and {Schelten}, Alan and {Yang}, Amy and Others},
        title = "{The Llama 3 Herd of Models}",
      journal = {arXiv e-prints},
         year = 2024,
        month = jul,
          eid = {arXiv:2407.21783},
        pages = {arXiv:2407.21783},
          doi = {10.48550/arXiv.2407.21783}
}

@ARTICLE{2024arXiv240512819Q,
       author = {{Qin}, Libo and {Chen}, Qiguang and {Feng}, Xiachong and {Wu}, Yang and {Zhang}, Yongheng and {Li}, Yinghui and {Li}, Min and {Che}, Wanxiang and {Yu}, Philip S.},
        title = "{Large Language Models Meet NLP: A Survey}",
      journal = {arXiv e-prints},
         year = 2024,
        month = may,
          eid = {arXiv:2405.12819},
        pages = {arXiv:2405.12819},
          doi = {10.48550/arXiv.2405.12819}
}

@article{Heersmink2024,
author={Heersmink, Richard},
title={Use of large language models might affect our cognitive skills},
journal={Nature Human Behaviour},
year={2024},
month={May},
day={01},
volume={8},
number={5},
pages={805-806},
issn={2397-3374},
doi={10.1038/s41562-024-01859-y},
url={https://doi.org/10.1038/s41562-024-01859-y}
}

@article{DARVISHI2024104967,
title = {Impact of AI assistance on student agency},
journal = {Computers \& Education},
volume = {210},
pages = {104967},
year = {2024},
issn = {0360-1315},
doi = {https://doi.org/10.1016/j.compedu.2023.104967},
url = {https://www.sciencedirect.com/science/article/pii/S0360131523002440},
author = {Ali Darvishi and Hassan Khosravi and Shazia Sadiq and Dragan Gašević and George Siemens}
}

@article{hume1996hinting,
 ISSN = {10508406, 15327809},
 URL = {http://www.jstor.org/stable/1466758},
 author = {Gregory Hume and Joel Michael and Allen Rovick and Martha Evens},
 journal = {The Journal of the Learning Sciences},
 number = {1},
 pages = {23--47},
 publisher = {Taylor & Francis, Ltd.},
 title = {Hinting as a Tactic in One-on-One Tutoring},
 urldate = {2024-08-19},
 volume = {5},
 year = {1996}
}

@ARTICLE{2024arXiv240404728J,
    title = "Navigating the Landscape of Hint Generation Research: From the Past to the Future",
    author = "Jangra, Anubhav  and
      Mozafari, Jamshid  and
      Jatowt, Adam  and
      Muresan, Smaranda",
    journal = "Transactions of the Association for Computational Linguistics",
    volume = "13",
    year = "2025",
    address = "Cambridge, MA",
    publisher = "MIT Press",
    url = "https://aclanthology.org/2025.tacl-1.25/",
    doi = "10.1162/tacl_a_00751",
    pages = "505--528"
}

@article{bandura2013role,
  title={The role of self-efficacy in goal-based motivation},
  author={Bandura, Albert},
  journal={New developments in goal setting and task performance},
  pages={147--157},
  year={2013},
  publisher={Routledge}
}

@inproceedings{jatowt_kg_hint,
author = {Jatowt, Adam and Gehrer, Calvin and F\"{a}rber, Michael},
title = {Automatic Hint Generation},
year = {2023},
isbn = {9798400700736},
publisher = {Association for Computing Machinery},
address = {New York, NY, USA},
url = {https://doi.org/10.1145/3578337.3605119},
doi = {10.1145/3578337.3605119},
booktitle = {Proceedings of the 2023 ACM SIGIR International Conference on Theory of Information Retrieval},
pages = {117-–123},
numpages = {7},
location = {Taipei, Taiwan},
series = {ICTIR '23}
}

@article{10.1145/3469885,
author = {McBroom, Jessica and Koprinska, Irena and Yacef, Kalina},
title = {A Survey of Automated Programming Hint Generation: The HINTS Framework},
year = {2021},
issue_date = {November 2022},
publisher = {Association for Computing Machinery},
address = {New York, NY, USA},
volume = {54},
number = {8},
issn = {0360-0300},
url = {https://doi.org/10.1145/3469885},
doi = {10.1145/3469885},
journal = {ACM Comput. Surv.},
month = oct,
articleno = {172},
numpages = {27}
}

@Article{Kochmar2022,
author={Kochmar, Ekaterina
and Vu, Dung Do
and Belfer, Robert
and Gupta, Varun
and Serban, Iulian Vlad
and Pineau, Joelle},
title={Automated Data-Driven Generation of Personalized Pedagogical Interventions in Intelligent Tutoring Systems},
journal={International Journal of Artificial Intelligence in Education},
year={2022},
month={Jun},
day={01},
volume={32},
number={2},
pages={323-349},
issn={1560-4306},
doi={10.1007/s40593-021-00267-x},
url={https://doi.org/10.1007/s40593-021-00267-x}
}

@Article{Price2019,
author={Price, Thomas W.
and Dong, Yihuan
and Zhi, Rui
and Paa{\ss}en, Benjamin
and Lytle, Nicholas
and Catet{\'e}, Veronica
and Barnes, Tiffany},
title={A Comparison of the Quality of Data-Driven Programming Hint Generation Algorithms},
journal={International Journal of Artificial Intelligence in Education},
year={2019},
month={Aug},
day={01},
volume={29},
number={3},
pages={368-395},
issn={1560-4306},
doi={10.1007/s40593-019-00177-z},
url={https://doi.org/10.1007/s40593-019-00177-z}
}

@inproceedings{mozafari-triviahg,
author = {Mozafari, Jamshid and Jangra, Anubhav and Jatowt, Adam},
title = {TriviaHG: A Dataset for Automatic Hint Generation from Factoid Questions},
year = {2024},
isbn = {9798400704314},
publisher = {Association for Computing Machinery},
address = {New York, NY, USA},
url = {https://doi.org/10.1145/3626772.3657855},
doi = {10.1145/3626772.3657855},
pages = {2060–-2070},
numpages = {11},
location = {Washington DC, USA},
series = {SIGIR '24}
}

@inproceedings{mozafari-hintqa,
    title = "Exploring Hint Generation Approaches for Open-Domain Question Answering",
    author = "Mozafari, Jamshid  and
      Abdallah, Abdelrahman  and
      Piryani, Bhawna  and
      Jatowt, Adam",
    editor = "Al-Onaizan, Yaser  and
      Bansal, Mohit  and
      Chen, Yun-Nung",
    booktitle = "Findings of the Association for Computational Linguistics: EMNLP 2024",
    month = nov,
    year = "2024",
    address = "Miami, Florida, USA",
    publisher = "Association for Computational Linguistics",
    url = "https://aclanthology.org/2024.findings-emnlp.546/",
    doi = "10.18653/v1/2024.findings-emnlp.546",
    pages = "9327--9352"
}

@inproceedings{joshi-etal-2017-triviaqa,
    title = "{T}rivia{QA}: A Large Scale Distantly Supervised Challenge Dataset for Reading Comprehension",
    author = "Joshi, Mandar  and
      Choi, Eunsol  and
      Weld, Daniel  and
      Zettlemoyer, Luke",
    editor = "Barzilay, Regina  and
      Kan, Min-Yen",
    booktitle = "Proceedings of the 55th Annual Meeting of the Association for Computational Linguistics (Volume 1: Long Papers)",
    month = jul,
    year = "2017",
    address = "Vancouver, Canada",
    publisher = "Association for Computational Linguistics",
    url = "https://aclanthology.org/P17-1147",
    doi = "10.18653/v1/P17-1147",
    pages = "1601--1611"
}

@article{kwiatkowski-etal-2019-natural,
    title = "Natural Questions: A Benchmark for Question Answering Research",
    author = "Kwiatkowski, Tom  and
      Palomaki, Jennimaria  and
      Redfield, Olivia  and
      Collins, Michael  and
      Parikh, Ankur  and
      Alberti, Chris  and
      Others",
    editor = "Lee, Lillian  and
      Johnson, Mark  and
      Roark, Brian  and
      Nenkova, Ani",
    journal = "Transactions of the Association for Computational Linguistics",
    volume = "7",
    year = "2019",
    address = "Cambridge, MA",
    publisher = "MIT Press",
    url = "https://aclanthology.org/Q19-1026",
    doi = "10.1162/tacl_a_00276",
    pages = "452--466"
}

@inproceedings{berant-etal-2013-semantic,
    title = "Semantic Parsing on {F}reebase from Question-Answer Pairs",
    author = "Berant, Jonathan  and
      Chou, Andrew  and
      Frostig, Roy  and
      Liang, Percy",
    editor = "Yarowsky, David  and
      Baldwin, Timothy  and
      Korhonen, Anna  and
      Livescu, Karen  and
      Bethard, Steven",
    booktitle = "Proceedings of the 2013 Conference on Empirical Methods in Natural Language Processing",
    month = oct,
    year = "2013",
    address = "Seattle, Washington, USA",
    publisher = "Association for Computational Linguistics",
    url = "https://aclanthology.org/D13-1160",
    pages = "1533--1544",
}

@inproceedings{10.1145/3404835.3463238,
author = {Lin, Jimmy and Ma, Xueguang and Lin, Sheng-Chieh and Yang, Jheng-Hong and Pradeep, Ronak and Nogueira, Rodrigo},
title = {Pyserini: A Python Toolkit for Reproducible Information Retrieval Research with Sparse and Dense Representations},
year = {2021},
isbn = {9781450380379},
publisher = {Association for Computing Machinery},
address = {New York, NY, USA},
url = {https://doi.org/10.1145/3404835.3463238},
doi = {10.1145/3404835.3463238},
booktitle = {Proceedings of the 44th International ACM SIGIR Conference on Research and Development in Information Retrieval},
pages = {2356-–2362},
numpages = {7},
location = {Virtual Event, Canada},
series = {SIGIR '21}
}

@inproceedings{ushio-etal-2023-practical,
    title = "A Practical Toolkit for Multilingual Question and Answer Generation",
    author = "Ushio, Asahi  and
      Alva-Manchego, Fernando  and
      Camacho-Collados, Jose",
    editor = "Bollegala, Danushka  and
      Huang, Ruihong  and
      Ritter, Alan",
    booktitle = "Proceedings of the 61st Annual Meeting of the Association for Computational Linguistics (Volume 3: System Demonstrations)",
    month = jul,
    year = "2023",
    address = "Toronto, Canada",
    publisher = "Association for Computational Linguistics",
    url = "https://aclanthology.org/2023.acl-demo.8",
    doi = "10.18653/v1/2023.acl-demo.8",
    pages = "86--94"
}

@inbook{10.5555/3454287.3455008,
author = {Paszke, Adam and Gross, Sam and Massa, Francisco and Lerer, Adam and Bradbury, James and Chanan, Gregory and Killeen, Trevor and Lin, Zeming and Gimelshein, Natalia and Antiga, Luca and Desmaison, Alban and K\"{o}pf, Andreas and Yang, Edward and DeVito, Zach and Raison, Martin and Tejani, Alykhan and Chilamkurthy, Sasank and Steiner, Benoit and Fang, Lu and Others},
title = {PyTorch: an imperative style, high-performance deep learning library},
year = {2019},
publisher = {Curran Associates Inc.},
address = {Red Hook, NY, USA},
booktitle = {Proceedings of the 33rd International Conference on Neural Information Processing Systems},
articleno = {721},
numpages = {12}
}

@inproceedings{2024arXiv241201626M,
    author = {Mozafari, Jamshid and Gerhold, Florian and Jatowt, Adam},
    title = {WikiHint: A Human-Annotated Dataset for Hint Ranking and Generation},
    year = {2025},
    isbn = {9798400715921},
    publisher = {Association for Computing Machinery},
    address = {New York, NY, USA},
    url = {https://doi.org/10.1145/3726302.3730284},
    doi = {10.1145/3726302.3730284},
    booktitle = {Proceedings of the 48th International ACM SIGIR Conference on Research and Development in Information Retrieval},
    pages = {3821-–3831},
    numpages = {11},
    location = {Padua, Italy},
    series = {SIGIR '25}
}

@Article{soc15010006,
AUTHOR = {Gerlich, Michael},
TITLE = {{AI Tools in Society: Impacts on Cognitive Offloading and the Future of Critical Thinking}},
JOURNAL = {Societies},
VOLUME = {15},
YEAR = {2025},
NUMBER = {1},
ARTICLE-NUMBER = {6},
URL = {https://www.mdpi.com/2075-4698/15/1/6},
ISSN = {2075-4698},
DOI = {10.3390/soc15010006}
}

@inproceedings{10.1145/3706598.3713778,
author = {Lee, Hao-Ping (Hank) and Sarkar, Advait and Tankelevitch, Lev and Drosos, Ian and Rintel, Sean and Banks, Richard and Wilson, Nicholas},
title = {{The Impact of Generative AI on Critical Thinking: Self-Reported Reductions in Cognitive Effort and Confidence Effects From a Survey of Knowledge Workers}},
year = {2025},
isbn = {9798400713941},
publisher = {Association for Computing Machinery},
address = {New York, NY, USA},
url = {https://doi.org/10.1145/3706598.3713778},
doi = {10.1145/3706598.3713778},
booktitle = {Proceedings of the 2025 CHI Conference on Human Factors in Computing Systems},
articleno = {1121},
numpages = {22},
location = {
},
series = {CHI '25}
}

@article{singh2025protecting,
  title={{Protecting human cognition in the age of AI}},
  author={Singh, Anjali and Taneja, Karan and Guan, Zhitong and Ghosh, Avijit},
  journal={arXiv preprint arXiv:2502.12447},
  year={2025}
}

@article{jangra2025designing,
  title={Designing and Evaluating Hint Generation Systems for Science Education},
  author={Jangra, Anubhav and Muresan, Smaranda},
  journal={arXiv preprint arXiv:2510.21087},
  year={2025}
}

@article{Honnibal_spaCy_Industrial-strength_Natural_2020,
author = {Honnibal, Matthew and Montani, Ines and Van Landeghem, Sofie and Boyd, Adriane},
doi = {10.5281/zenodo.1212303},
title = {{spaCy: Industrial-strength Natural Language Processing in Python}},
year = {2020}
}

@inproceedings{mozafari2026inferentialqa,
author = {Mozafari, Jamshid and Zamani, Hamed and Zuccon, Guido and Jatowt, Adam},
title = {Inferential Question Answering},
year = {2026},
isbn = {9798400723070},
publisher = {Association for Computing Machinery},
address = {New York, NY, USA},
url = {https://doi.org/10.1145/3774904.3792653},
doi = {10.1145/3774904.3792653},
booktitle = {Proceedings of the ACM Web Conference 2026},
pages = {2384-–2395},
numpages = {12},
location = {United Arab Emirates},
series = {WWW '26}
}

@inproceedings{mozafari2026context,
author = {Mozafari, Jamshid and Piryani, Bhawna and Jatowt, Adam},
title = {Context Convergence Improves Answering Inferential Questions},
year = {2026},
isbn = {9798400725999},
publisher = {Association for Computing Machinery},
address = {New York, NY, USA},
url = {https://doi.org/10.1145/3805712.3809848},
doi = {10.1145/3805712.3809848},
booktitle = {Proceedings of the 49th International ACM SIGIR Conference on Research and Development in Information Retrieval},
pages = {4010-–4015},
numpages = {6},
location = {Australia},
series = {SIGIR '26}
}
